\documentclass{article}

\usepackage[final]{neurips_2025}

\usepackage[utf8]{inputenc} 
\usepackage[T1]{fontenc}    
\usepackage{hyperref}       
\usepackage{url}            
\usepackage{booktabs}       
\usepackage{amsfonts}       
\usepackage{algorithm}
\usepackage{algpseudocode}
\usepackage{enumitem}
\usepackage{wrapfig}
\usepackage{amsmath}
\usepackage{amssymb}
\usepackage{bm}
\usepackage{svg}
\usepackage{nicefrac}       
\usepackage{microtype}      
\usepackage{xcolor}         
\usepackage{multirow}
\usepackage{graphicx}
\usepackage{subcaption}
\usepackage{cleveref}
\usepackage{fancyvrb}
\usepackage{xcolor}
\usepackage{pgfplots}
\pgfplotsset{compat=1.18}
\usepackage{threeparttable}
\usepackage{tikz}
\usetikzlibrary{positioning,calc}

\usepackage{titlesec}
\titlespacing*{\paragraph}{0pt}{0.5ex}{0.5ex}

\title{Symphony for Speech-to-Text: \\Supporting Real-Time Medical Voice Interfaces}

\author{%
  Arne Nix \\
  \And
  Robert James \\
  \And
  Lasse Borgholt \\
  \And
  Anna B. Ekner \\
  \And
  Lana Krumm \\
  \And
  Julius Severin \\
  \And
  Dan Engel \\
  \And
  Lars Maal{\o}e \\
  \And
  Jakob Havtorn* \\
  \AND
  Corti \\
}

\begin{document}

\maketitle

\begin{abstract}
After decades of use in dictation and, more recently, ambient documentation, speech is emerging as a primary modality for interacting with technology and AI in healthcare. Yet medical speech recognition remains difficult: systems must capture specialized terminology, resolve contextual ambiguity, and render measurements, abbreviations, and clinical shorthand precisely. Existing solutions are typically optimized either for general-purpose transcription or narrow dictation workflows, limiting their reliability in safety-critical settings and their usefulness for broader clinical workflows. We introduce Symphony for Speech-to-Text, a medical-grade speech recognition system for real-time streaming and batch file-based clinical use. Symphony decomposes the transcription process into specialized components for recognition, formatting, and contextual correction to optimize medical term recall while producing clinically structured text in real time and adapting across use cases. Evaluations on public benchmark and medical speech datasets show that Symphony substantially outperforms state-of-the-art systems in clinical settings while matching or exceeding them in general-domain settings, suggesting robust generalization rather than overfitting. We release a clinical benchmark dataset to support reliable validation and further progress in medical speech recognition. Symphony is available through a production-grade API for live dictation, conversational transcription, and batch audio file processing.
\end{abstract}

\section{Introduction}
Speech is becoming a core interface for healthcare. Clinicians already use speech recognition for documentation capture, electronic health record (EHR) navigation, and workflow automation, and the next generation of clinical software will increasingly rely on speech-enabled interaction for ambient workflows, structured data capture, and real-time assistance \citep{Goss_2019, Falcetta_2023, Haberle_2024}. In this setting, speech recognition is not merely a transcription utility; it is a foundational layer for both the everyday software clinicians rely on and the AI-enabled solutions healthcare builders are advancing.

This raises a higher bar than general-purpose automatic speech recognition is designed to meet. In clinical settings, systems must accurately capture rare and specialized terminology, preserve distinctions that are medically consequential, and render shorthand, acronyms, dosages, and measurements in a form that is immediately usable downstream \citep{Hodgson_2017, Zhou_2018}. Errors at this level are not only cosmetic. They increase correction burden, degrade trust, and limit the usefulness of speech interfaces in real-world care delivery.

Most existing speech recognition systems are trained and evaluated as broad-domain transcription models \citep{Radford_2023}. While they have improved substantially in recent years, their core objective remains the production of plausible text from audio. That objective is insufficient for medicine, where correctness depends not only on word recognition, but also on domain-specific formatting, contextual disambiguation, and adaptation to the task at hand. A system that transcribes “fifty” correctly but fails to render a dosage, unit, acronym, or abbreviation in its clinically intended form is often still wrong in practice. Conversely, systems optimized narrowly for single-speaker medical dictation often fail to generalize to conversational clinical audio, where overlapping speech, turn-taking, disfluencies, and variable acoustic conditions make transcription substantially harder \citep{Liesenfeld_2023}. In these settings, such systems may not only miss clinically relevant content, but also produce unstable outputs, including ungrounded insertions and other hallucinated text, further limiting their reliability in real-world use \citep{frieske2024hallucinationsneuralautomaticspeech}.

Recent work has begun to use large language models to improve transcription outputs through rescoring, reformatting, or post-hoc correction \citep{Yang_2023, radhakrishnan_2023}. These methods show that medical speech recognition benefits from stronger language understanding, but are often introduced as loosely coupled post-processing steps rather than as an integrated system design \citep{hu_2024}. As a result, they do not fully address the need for real-time operation, structured output, and task-specific adaptation across different clinical workflows.

Therefore, we argue that medical speech recognition should not be viewed as a single-stage transcription problem. Rather, it is better understood as a structured inference problem in which raw recognition, linguistic normalization, and contextual correction must work together to produce text that is clinically faithful and operationally useful.

We present Symphony for Speech-to-Text, combining a streaming recognition model specialized for high-precision and high-recall capture of medical terminology with support for keyterm biasing, a real-time language model for rendering structured clinical text, and a correction model that uses contextual reasoning and custom prompts to adapt outputs to different settings. The system also exposes real-time audio-quality signals that surface input audio quality problems. The result is a system designed not only to accurately transcribe speech but also to serve as an infrastructure for speech-enabled clinical workflows.

In this paper, we make the following contributions:
\begin{enumerate}
    \item \textbf{Keyterm precision and recall:} We introduce a real-time speech recognition model specialized for the accurate recognition of medical terminology, spoken-punctuation symbols, and formatted entities from clinical audio.
    
    \item \textbf{Structured transcription:} We introduce a real-time language model that formats acronyms, measurements, and other clinically important textual conventions during transcription.
    
    \item \textbf{Contextual correction:} We present a language model for context-aware correction that can be adapted through custom prompts to different downstream use cases and documentation styles.
    
    \item \textbf{Comprehensive evaluation:} We show that Symphony performs on par with or better than best-in-class systems on public benchmarks, while substantially outperforming competing systems on medical speech recognition datasets. We adopt a keyterm precision and recall evaluation that separately measures medical terms, formatted entities, and spoken punctuation, exposing failure modes that aggregate or recall-only metrics obscure.

    \item \textbf{Dataset release:} We release a new medical speech recognition dataset to support further improvement and more realistic evaluation across the field.
    
    \item \textbf{Deployment across workflows:} We demonstrate a single production system that supports dictation, conversational speech, and batch audio file processing through a unified API\footnote{Find more information on \href{https://docs.corti.ai/stt}{docs.corti.ai/stt} and try a demo on \href{https://console.corti.app}{console.corti.app}.}.
    
    \item \textbf{Audio quality awareness:} We provide real-time audio events that expose input-quality conditions, enabling more reliable operation and better monitoring of speech-based systems.
\end{enumerate}

\section{Methods}
\subsection{Symphony for speech-to-text}
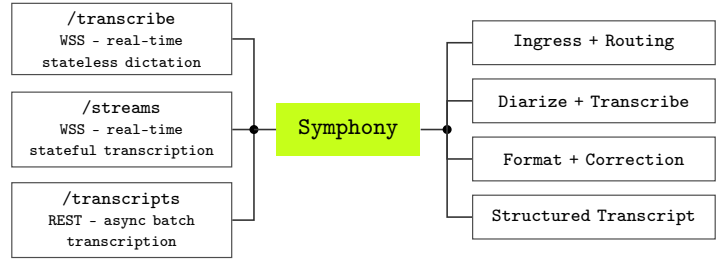
\begin{figure}[t]
\centering
\begin{tikzpicture}[
    font=\ttfamily\scriptsize,
    block/.style={
        rectangle,
        draw=black!70,
        line width=0.45pt,
        fill=white,
        align=center,
        minimum height=0.58cm,
        text width=2.7cm,
        inner sep=3pt
    },
    core/.style={
        rectangle,
        fill=lime!90,
        draw=none,
        align=center,
        minimum width=1.9cm,
        minimum height=0.7cm,
        font=\ttfamily\small
    },
    conn/.style={draw=black!80, line width=0.6pt},
    dot/.style={circle, fill=black, inner sep=1.2pt}
]

\node[block] (e1) at (0,  1.2) {\texttt{/transcribe \\ \tiny{WSS - real-time stateless dictation}}};
\node[block] (e2) at (0,  0.0) {\texttt{/streams \\ \tiny{WSS - real-time stateful transcription}}};
\node[block] (e3) at (0, -1.2) {\texttt{/transcripts \tiny{REST - async batch transcription}}};

\node[core] (s) at (3.0, 0) {Symphony};

\node[block, text width=3.0cm] (p1) at (6.25,  1.15) {Ingress + Routing};
\node[block, text width=3.0cm] (p2) at (6.25,  0.38) {Diarize + Transcribe};
\node[block, text width=3.0cm] (p3) at (6.25, -0.39) {Format + Correction};
\node[block, text width=3.0cm] (p4) at (6.25, -1.16) {Structured Transcript};

\coordinate (j1) at (1.75, 0);

\draw[conn] (e1.east) -| (j1);
\draw[conn] (e2.east) -- (j1);
\draw[conn] (e3.east) -| (j1);
\node[dot] at (j1) {};

\draw[conn] (j1) -- (s.west);

\coordinate (stem) at ($(s.east)+(0.35,0)$);
\draw[conn] (s.east) -- (stem);
\node[dot] at (stem) {};

\draw[conn] (stem) |- (p1.west);
\draw[conn] (stem) |- (p2.west);
\draw[conn] (stem) |- (p3.west);
\draw[conn] (stem) |- (p4.west);

\end{tikzpicture}
\caption{Symphony API endpoints and shared speech-to-text pipeline. The system supports stateless real-time dictation through \texttt{/transcribe}, stateful real-time transcription through \texttt{/streams}, and asynchronous batch transcription through \texttt{/transcripts}. All endpoints feed into a common processing pipeline that performs audio ingress and routing, diarization and transcription, formatting and contextual correction, and generation of structured transcripts.}
\label{fig:symphony-api-pipeline}
\end{figure}

\paragraph{API endpoints}
Symphony exposes speech recognition through three API endpoints that correspond to the main ways clinical applications interact with audio (Figure \ref{fig:symphony-api-pipeline}). The \texttt{/transcribe} endpoint supports real-time, stateless dictation over WebSocket Secure (WSS), making it suitable for command-and-control dictation workflows and speech-enabled user interfaces. The \texttt{/streams} endpoint provides a real-time, stateful WSS interface for conversational transcription and clinical intelligence, where audio is associated with an ongoing interaction. Finally, \texttt{/transcripts} supports asynchronous transcription over REST for batch processing of pre-recorded audio files. Together, these endpoints allow the same underlying speech-to-text system to support live dictation and ambient conversations or audio-file-based workflows through different integration patterns.

\paragraph{Ingress and routing}
Audio entering Symphony is first normalized through an ingress and routing layer. This layer abstracts the differences between live WebSocket audio and uploaded recordings, allowing downstream components to operate on a common representation of the audio stream. In the streaming case, the system returns low-latency interim and final recognition results to support responsive user interfaces; in the batch case, uploaded recordings are processed through an asynchronous transcript creation workflow. This separation between endpoint semantics and recognition infrastructure allows Symphony to provide different interface behaviors for integrators while sharing a common transcription pipeline.

\paragraph{Recognition and diarization}
The central recognition stage performs medical speech recognition and, when configured, speaker diarization. The recognition model is optimized for high-precision and high-recall capture of clinical terminology, including medications, diagnoses, procedures, measurements, and abbreviations. For multi-speaker clinical conversations, diarization segments the transcript by speaker and assigns speech to distinct participants, improving readability and reviewability of conversational transcripts. In multichannel workflows, speaker attribution can instead be derived from channel configuration, which provides exact attribution when separate audio channels are available per speaker.

\paragraph{Formatting and contextual correction}
After recognition, Symphony applies formatting, based on entity tagging, and contextual correction. This stage converts spoken clinical language into text that is directly useful for documentation and downstream workflows while staying grounded in the verbatim transcript. Formatting handles domain-relevant surface forms such as dates, times, numbers, measurements, numeric ranges, and ordinals, all of which are frequent sources of errors in clinical speech recognition. For example, spoken measurements and numeric expressions must often be rendered in standardized textual forms to be safe and useful in clinical documentation.  

Symphony also incorporates contextual reasoning and configurable prompting to adapt outputs to different clinical settings. This is important because the correct rendering of speech often depends on local context: the same acoustic signal may correspond to different abbreviations, medication names, units, or documentation conventions depending on specialty, workflow, and surrounding utterances. By separating recognition, formatting, and correction into distinct components, Symphony can improve raw transcription accuracy while also producing text that is configurable to match the exact expectations of different clinical applications.

\paragraph{Audio-quality feedback}
In real-time deployments, the system also exposes audio-quality events to the client. These events allow applications to surface problems with speech intelligibility or unexpected, prolonged silence while the interaction is still in progress. This feedback loop is important for clinical voice interfaces because recognition quality depends not only on the model but also on recording conditions including microphone placement and configuration, user behavior, and noise types and levels. Unlike other consumer applications, a clinical encounter is a singular event where audio lost to poor conditions is unrecoverable. By making audio health visible to the application, Symphony supports interfaces that can prompt the user to correct input conditions before transcript quality is compromised, or the opportunity is lost.

\paragraph{Command and control}
In addition to transcription, Symphony supports command-and-control workflows through its real-time API. The \texttt{/transcribe} endpoint is designed for stateless dictation and speech-enabled applications, where spoken commands can be used to control the user interface, edit text, trigger formatting operations, or support hands-free interaction. This makes the API suitable not only for converting speech into text, but also for building speech-enabled clinical interfaces in which speech acts as both an input modality for documentation and a control modality for interacting with electronic systems.

\paragraph{Training}
We train Symphony on a mixture of publicly available speech and text-to-speech data, together with a large corpus of proprietary data collected by Corti. The training corpus spans a wide range of real-world acoustic and linguistic conditions, including medical consultations, clinical dictations, conversational speech, and everyday interactions. To improve robustness and coverage, we combine real recordings with synthetically generated examples that increase the representation of rare medical terminology, abbreviations, medications, measurements, and domain-specific phrasing. This mixture exposes the model to both natural variability in spoken language and targeted clinical edge cases, helping it generalize across speakers, accents, recording conditions, and workflow types.

Overall, Symphony is designed as an API-level infrastructure for speech-enabled clinical workflows, not merely a transcription model. Endpoints define how applications submit audio, the shared pipeline performs medical recognition and speaker-aware transcription, and the post-processing layer transforms recognized speech into structured, clinically usable text and commands.

\subsection{Evaluation data}
We evaluate Symphony on datasets that represent four complementary capabilities required for medical speech recognition: general clinical dictation, radiology dictation, medical terminology coverage, and general-domain robustness. 

Although Symphony is available across many languages, we focus our evaluation on English, German, and French. These languages provide a controlled multilingual benchmark spanning different pronunciation patterns, morphology, abbreviation conventions, and clinical documentation styles. This allows us to evaluate whether Symphony generalizes beyond English while maintaining high-quality annotations and comparable evaluation sets across systems. The language scope of the benchmark should therefore not be interpreted as the scope of the deployed system: Symphony supports a broader set of languages through the Corti API, with additional languages available upon request.

\begin{table}[t]
\centering
\caption{Dataset statistics for speech recognition evaluation sets.}
\vspace{0.25cm}
\begin{tabular}{llrrrrr}
\toprule
\textbf{Dataset} &  \textbf{Lang} & \textbf{Examples} & \textbf{Duration} & \textbf{Med. terms} & \textbf{Fmt. terms} \\
\midrule
\multirow{3}{*}{MedDictate}  & EN & 24  & 65.8 min & 844  & 272  \\
                              & FR & 7   & 15.5 min & 169   & 34   \\
                              & DE & 9   & 32.6 min & 488   & 145  \\
\midrule
\multirow{3}{*}{MedTerm}     & EN & 499 & 9.9 h & 1,435  & 1,404  \\
                              & FR & 457 & 9.7 h & 1,346  & 1,042  \\
                              & DE & 499 & 9.4 h & 1,402  & 1,080  \\
\midrule
\multirow{1}{*}{MedRad}     & EN & 5000 & 30 h & 1,920  & -  \\
\midrule
\multirow{3}{*}{CommonVoice}& EN & 2,000 & 3.29 h & - & - \\
                            & FR & 2,000 & 3.27 h & - & - \\
                            & DE & 2,000 & 3.45 h & - & - \\
\bottomrule
\end{tabular}
\label{tab:dataset-stats}
\end{table}

\paragraph{MedDictate}
MedDictate\footnote{We will make MedDictate available on Hugging Face at:  \href{https://huggingface.co/datasets/corti/med-dictate}{\texttt{corti/med-dictate}}.} is a carefully curated medical dictation dataset in English, French, and German dictated by medical professionals, totalling nearly two hours of audio with gold-standard transcripts. The dataset spans a broad range of clinical domains, from radiology to psychology, and is designed to reflect realistic documentation scenarios. The underlying notes were hand-crafted to include challenging clinical language, diverse dictation styles, and terminology-rich content, encompassing a broad set of unique medical terms. MedDictate therefore evaluates end-to-end dictation quality, including recognition of medical terminology, preservation of clinically relevant details, and accurate rendering of dictated clinical text. 

\paragraph{MedTerm}
MedTerm is a large-scale dataset of short medical dictations, each constructed around specialized medical terms uniformly sampled from a terminology database containing more than 100,000 terms per language. The dataset is designed to stress-test medical vocabulary coverage across English, French, and German, with particular emphasis on rare, specialized, and safety-critical terminology that may be underrepresented in general-purpose speech recognition benchmarks. In addition to medical terms, MedTerm includes formatted entities and spoken punctuation, enabling evaluation of both domain-specific recognition and clinically important text normalization.

\paragraph{MedRad}
MedRad\footnote{We only use MedRad for Symphony keyterm biasing evaluations due to PHI.} is an English radiology dictation dataset consisting of a large set of samples. Radiology is a key evaluation domain for medical speech recognition because it is one of the most established and high-volume uses of clinical dictation. Radiology reports are terminology-dense and frequently contain measurements, anatomical references, laterality, negation, and concise diagnostic phrasing. Errors in these elements can materially change the interpretation of a report, making radiology a strong stress test for precision, consistency, and clinical usability in real-world dictation workflows.

\paragraph{Common Voice}
Common Voice\footnote{We use \href{https://commonvoice.mozilla.org/}{Common Voice} for general-domain speech recognition evaluation.} is a large, open, community-curated speech dataset designed to expand coverage across languages, accents, and speaker populations. Unlike more traditional English-only benchmarks, Common Voice reflects the diversity and heterogeneity of general, real-world speech contributed by volunteers. In our evaluation, we include Common Voice as a general-purpose benchmark to verify that Symphony's medical-domain specialization does not come at the cost of regression on everyday speech recognition.

\paragraph{FSD50K}
FSD50K\footnote{We use \href{https://annotator.freesound.org/fsd/release/FSD50K/}{FSD50K} for controlled background-noise robustness experiments.} is a large-scale collection of everyday sound events used to evaluate robustness under realistic background-noise conditions. We use samples from FSD50K for non-speech acoustic interference and mix them with speech at different signal-to-noise ratios. This allows us to test whether Symphony's audio-quality events respond to degraded input conditions and whether those events correlate with downstream transcription quality. Unlike the speech datasets above, FSD50K is not used to directly evaluate transcription accuracy, but to create controlled noisy audio conditions to assess the detection of audio-health events.

\section{Experiments}

\subsection{Evaluation metrics}

\paragraph{Word error rate (WER)}
WER measures the edit distance between a reference transcript and a hypothesis transcript at the word level, normalized by the length of the reference. Given a reference of $N$ words, let $S$, $D$, and $I$ denote the minimum number of word-level substitutions, deletions, and insertions required to transform the hypothesis into the reference under Levenshtein alignment. WER is defined as
\begin{equation}
    \mathrm{WER} = \frac{S + D + I}{N}.
\end{equation}
Lower WER indicates better performance. Because WER counts substitutions, deletions, and insertions relative to the reference transcript, it can exceed 100\% when a system adds many extra words. WER should only be compared when the reference and model output are normalized in the same way, for example with consistent casing, punctuation, and number formatting.

\paragraph{Keyterm precision and recall}
For word categories that are clinically or functionally important, such as medical entities, formatted entities, and spoken punctuation, we report a standalone recall for each category ($R_{\mathrm{med}}, R_{\mathrm{fmt}}, R_{\mathrm{punct}}$). For spoken punctuation, we additionally report precision ($P_{\mathrm{punct}}$), since many systems produce unwarranted punctuation in dictation settings, where they are expected to insert punctuation only in response to explicit spoken commands. This behavior can cause a sharp rise in false positives that recall alone does not capture.

Let $\mathcal{V}$ be a keyterm vocabulary, and let $\mathcal{A}$ be the word-level Levenshtein alignment between reference and hypothesis used for WER. Restricting attention to terms in $\mathcal{V}$:
\begin{itemize}[leftmargin=1.5em]
    \item $\mathrm{TP}$: the number of $\mathcal{V}$-terms in the reference whose aligned hypothesis tokens under $\mathcal{A}$ correspond to the same $\mathcal{V}$-term.
    \item $\mathrm{FN}$: the number of $\mathcal{V}$-terms in the reference whose aligned hypothesis tokens are missing or are not the same $\mathcal{V}$-token (recall failures).
    \item $\mathrm{FP}$: the number of $\mathcal{V}$-terms in the hypothesis whose aligned reference tokens are not the same $\mathcal{V}$-token (precision failures).
\end{itemize}
Precision and recall are then
\begin{equation}
    P_{\mathcal{V}} = \frac{\mathrm{TP}}{\mathrm{TP} + \mathrm{FP}}, \qquad R_{\mathcal{V}} = \frac{\mathrm{TP}}{\mathrm{TP} + \mathrm{FN}}.
\end{equation}
The disambiguation cases fall out of the definition: if the speaker uttered the literal word ``colon'' in an anatomical context and the system emitted ``:'', the symbol contributes an $\mathrm{FP}$ to $P_{\mathrm{punct}}$; if the symbol was intended and the system emitted the word, that contributes an $\mathrm{FN}$ to $R_{\mathrm{punct}}$. Reporting precision and recall separately, rather than aggregating into $F_1$ or a single error rate, exposes which side of the trade-off a system fails on.

\paragraph{Keyterm biasing ($\mathrm{rFNR}$, $P_{\mathrm{med}}$)}
We measure biasing effectiveness by the relative reduction in miss rate when a biasing vocabulary is provided, and monitor precision to verify that biasing does not introduce spurious detections. The relative FNR reduction under biasing is
\begin{equation}
    \mathrm{rFNR}_{\mathcal{V}} = \frac{\mathrm{FNR}_{\mathcal{V}}^{\mathrm{unbiased}} - \mathrm{FNR}_{\mathcal{V}}^{\mathrm{biased}}}{\mathrm{FNR}_{\mathcal{V}}^{\mathrm{unbiased}}},
\end{equation}
where $\mathrm{FNR}_{\mathcal{V}} = 1 - R_{\mathcal{V}}$. This quantity is defined whenever the unbiased system makes at least one miss. A value of $\mathrm{rFNR} = 0.5$ means biasing eliminates half of the unbiased system's misses; because the reduction is relative, it rewards the same proportional improvement whether baseline recall is 80\% or 98\%. To verify that recall gains do not come at the cost of spurious detections, we report $P_{\mathrm{med}}$ under both unbiased and biased conditions. Safe biasing leaves precision essentially unchanged; a notable drop in $P_{\mathrm{med}}^{\mathrm{biased}}$ relative to $P_{\mathrm{med}}^{\mathrm{unbiased}}$ would signal over-aggressive biasing. Defined analogously to $\mathrm{FNR}_{\mathcal{V}}$, we will also briefly use the false discovery rate, $\mathrm{FDR}_{\mathcal{V}} = 1 - P_{\mathcal{V}}$, which measures the fraction of predicted medical keyterms that are spurious.

\paragraph{Confidence estimation.} For each system, dataset, and metric, we estimate uncertainty using the percentile bootstrap. We draw $S=1000$ bootstrap samples with replacement at the utterance level from the evaluation set and recompute the metric on each resample. The metric computed on the full evaluation set is used as the point estimate, and the 2.5th and 97.5th percentiles of the bootstrap distribution define the 95\% confidence interval. This captures both sampling variability in the metric estimate and the sensitivity of system comparisons to the composition of the evaluation set, which is particularly important for smaller medical benchmarks.

\paragraph{Tooling}
All the evaluations conducted in this work are scored using BeWER\footnote{BeWER metrics library available at: \href{https://pypi.org/project/bewer/}{\texttt{https://pypi.org/project/bewer/}}.}, an open-source library developed and released by Corti. 
BeWER implements standard metrics (WER, CER), as well as the keyterm precision and recall metrics, percentile bootstrap confidence intervals, and normalization pipeline defined in the previous section.
For a qualitative analysis, Corti Canal\footnote{Corti Canal reporting tool available at: \href{https://pypi.org/project/corti-canal/}{\texttt{https://pypi.org/project/corti-canal}}.} wraps alignment and metric computation into an open-source command-line tool that produces self-contained visual reports with per-word error overlays and aggregate accuracy breakdowns. 
For internal metrics and analysis, we have recently adopted the string alignment method ErrorAlign\footnote{ErrorAlign string alignment algorithm available at \href{https://pypi.org/project/error-align/}{\texttt{https://pypi.org/project/error-align/}}} \citep{borgholt_2026} but use the standard Levenshtein alignment in this paper for comparability to most public results.

\subsection{Text normalization}

\paragraph{Formatting.}
Our goal is to identify the strongest overall transcription system for medical use among the systems we compare. Because accurate formatting of dosages, units, dates, and similar entities is essential in medical speech recognition, we compute all metrics against formatted reference transcripts that represent the expected output of a medical transcription system. 

Specifically, units follow SI abbreviations (e.g., "mg", "mL", "mmHg"), dates are rendered in standard long form for the respective locale (e.g., "February 3, 2025" in English, "3. Februar 2025" in German, "3 février 2025" in French), times follow locale convention (e.g., "8:00 AM" in English, "08:00" in German, "8h00" in French), cardinal numbers and ordinals from one through nine are written as words while values of 10 and above are rendered as numerals, and any numeric range is rendered with numerals (e.g., "4–15"). 

For general-domain transcription, we apply the same principle when formattable entities are present, although such entities occur less frequently.

\paragraph{Casing and punctuation.}
All metrics are computed case-insensitively. 
For WER, punctuation symbols are additionally stripped from both reference and hypothesis to be consistent with standard public benchmarks. 
The keyterm metrics $(P_\mathcal{V}, R_\mathcal{V})$ retain punctuation symbols, since spoken-punctuation handling is part of what they measure.
Hence, the reference contains the punctuation symbols corresponding to spoken punctuations command in the audio. 
This means punctuation is only expected when it was actually spoken. The spoken-punctuation vocabulary covers period (.), comma (,), new line, new paragraph, exclamation mark (!), question mark (?), colon (:), semicolon (;), hyphen (-), slash (/), quotation marks (" "), and parentheses ( ).

Concretely, any missed punctuation command will decrease $R_{\mathrm{punct}}$ and any punctuation symbol in the hypothesis without corresponding command in the audio will be penalized in $P_{\mathrm{punct}}$.
For example, the word ``colon'' in its anatomical sense should remain the word ``colon'' in the formatted reference; but ``colon'' meant as a dictation command should become ``:'' in the formatted reference. 
This makes spoken-punctuation evaluation an explicit disambiguation test where contextual understanding of the transcript is required for correct formatting. 

\subsection{Speech-to-text systems for comparison}

\begin{figure}[!t]
    \centering
    \href{http://corti.ai/compare/speech-to-text}{
        \includegraphics[width=0.97\linewidth]{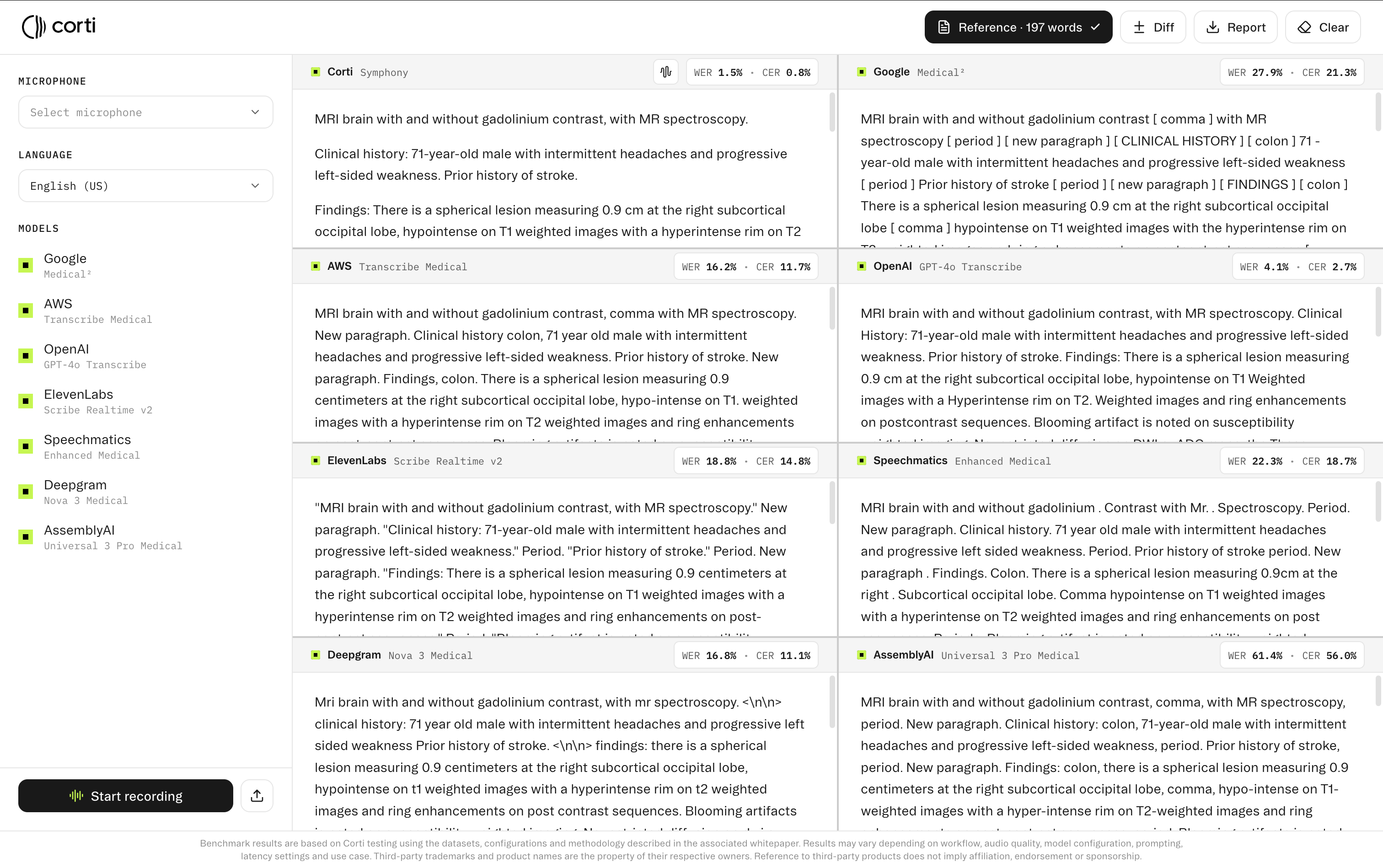}
    }
    \caption{Corti's speech-to-text comparison tool. The interactive tool allows users to compare transcription outputs across speech recognition systems on the same audio input. Available at \href{http://corti.ai/compare/speech-to-text}{corti.ai/compare/speech-to-text}.}
    \label{fig:corti_compare_tool}
\end{figure}

We compare Symphony against three classes of baseline: general-purpose cloud speech-to-text APIs, a dedicated medical dictation system (Dragon Medical One), and leading open-source models. This selection enables us test Symphony against both the strongest commercially available APIs and the specialized tooling already deployed in clinical documentation workflows.

We first conducted a pre-evaluation of major cloud-based speech recognition APIs, including Deepgram, AssemblyAI, Speechmatics, OpenAI, ElevenLabs, Google Cloud, Amazon Transcribe, and Microsoft Azure (Figure \ref{fig:corti_compare_tool}). These systems represent a broad cross-section of commercially available speech recognition services, including both batch and real-time APIs, and are commonly used as general-purpose transcription backends.

The pre-evaluation was used to identify the strongest general-purpose systems for more detailed comparison. 
Where a cloud API offered a medical-specific mode, we used it (e.g.\ Amazon Transcribe Medical, Google Cloud Medical). Among all systems considered, ElevenLabs, Amazon Transcribe Medical, Google Cloud, and OpenAI performed best in our initial screening and were selected for a full offline evaluation on our medical and general-domain benchmarks.

In our offline evaluation, OpenAI Speech-to-Text\footnote{While testing the OpenAI Speech-to-Text real-time system, we learned that the system relies on a conservative voice activity detection (VAD), making the time to receive a transcript very slow. There is a discussion to be had on whether this system can in fact be treated as a real-time speech-to-text system. In this paper we treat it as such for simplicity.} 
and ElevenLabs Speech-to-Text were the strongest performing general-purpose APIs overall, and were consequently selected for the real-time evaluation against Symphony. This two-stage procedure lets us screen broadly while focusing the real-time experiments on the strongest baselines.

We evaluate Dragon Medical One (DMO), a dedicated medical speech recognition product designed for clinical dictation. Unlike the cloud APIs above, DMO is not a general-purpose transcription backend but a clinician-facing documentation tool widely deployed in healthcare settings. Including DMO let us test whether Symphony improves not only over general-purpose APIs, but also over a specialized system that many clinicians already use for day-to-day documentation.

We also evaluate two leading open-source speech recognition architectures: Whisper and Parakeet. These models provide important reference points because they are widely used as research and deployment baselines outside proprietary cloud APIs. Whisper is a multilingual encoder-decoder model trained on large-scale weak supervision and has become a standard open-source baseline for robust general-purpose transcription \citep{radford2022whisper}. Parakeet is a family of high-performance ASR models released through NVIDIA NeMo, based on modern transducer-style architectures designed for accurate and efficient speech recognition \citep{nvidia2024parakeet,sekoyan2025canaryparakeet}. 

\section{Results}
We evaluate Symphony across four complementary settings: medical terminology coverage, realistic medical dictation, general-domain speech recognition, and keyterm biasing. The medical evaluations test whether systems can recognize specialized clinical vocabulary, render formatted entities, and handle spoken punctuation in dictation-style workflows. The CommonVoice evaluation serves as a general-domain control to assess whether medical specialization degrades broader speech recognition performance. Finally, the keyterm biasing experiment measures whether Symphony can safely adapt to user- or workflow-specific terminology without sacrificing precision.

Across these settings, we compare Symphony against leading open-source ASR models, commercial speech-to-text APIs, and, where applicable, a dedicated medical dictation system. We report both aggregated transcription quality through WER and clinically targeted metrics for medical terms, formatted entities, and spoken punctuation.

\subsection{Medical terminology}
\begin{table}[t]
    \centering
    \caption{\textbf{Real-time on MedTerm [English].} Comparison across WER and keyterm precision/recall ($P_{\mathcal{V}}, R_{\mathcal{V}}$) for medical entities ($\mathcal{V}=\mathrm{med}$), spoken-punctuation symbols ($\mathcal{V}=\mathrm{punct}$), and formatted entities ($\mathcal{V}=\mathrm{fmt}$). Higher is better for $P/R$ metrics; lower is better for WER. All values reported in percent $[\%]$. 95\% confidence intervals shown below each value. \textbf{Bold} indicates best in each row.}
    \vspace{0.5cm}
    \label{tab:med_term_en_realtime}
    \small
    \begin{tabular}{llccccc}
        \toprule
        & & \multicolumn{2}{c}{\textbf{Open-source}} & \multicolumn{2}{c}{\textbf{Closed-source}} & \textbf{Corti} \\
        \cmidrule(lr){3-4} \cmidrule(lr){5-6} \cmidrule(lr){7-7}
        \textbf{Lang} & \textbf{Metric}
        & \rotatebox{90}{\textbf{Parakeet}}
        & \rotatebox{90}{\textbf{Whisper}}
        & \rotatebox{90}{\textbf{OpenAI}}
        & \rotatebox{90}{\textbf{ElevenLabs}}
        & \rotatebox{90}{\textbf{Symphony}} \\
        \midrule
        & WER
        & \shortstack{18.9\\{\tiny{[18.4,19.5]}}}
        & \shortstack{17.4\\{\tiny{[16.9,17.9]}}}
        & \shortstack{17.7\\{\tiny{[17.1,18.3]}}}
        & \shortstack{18.1\\{\tiny{[17.5,18.7]}}}
        & \shortstack{$\mathbf{2.1}$\\{\tiny{[2.0,2.3]}}} \\[4pt]
        
        & $R_{\mathrm{med}}$
        & \shortstack{66.2\\{\tiny{[63.7,68.4]}}}
        & \shortstack{70.8\\{\tiny{[68.3,73.0]}}}
        & \shortstack{75.5\\{\tiny{[73.2,77.7]}}}
        & \shortstack{72.2\\{\tiny{[69.8,74.5]}}}
        & \shortstack{$\mathbf{81.8}$\\{\tiny{[80.0,83.7]}}} \\[4pt]
        
     EN & $R_{\mathrm{fmt}}$
        & \shortstack{37.6\\{\tiny{[34.1,41.1]}}}
        & \shortstack{44.3\\{\tiny{[40.9,47.7]}}}
        & \shortstack{33.4\\{\tiny{[29.5,37.1]}}}
        & \shortstack{43.1\\{\tiny{[39.7,46.8]}}}
        & \shortstack{$\mathbf{93.0}$\\{\tiny{[91.5,94.3]}}} \\[4pt]
        
        & $R_{\mathrm{punct}}$
        & \shortstack{41.1\\{\tiny{[39.6,42.5]}}}
        & \shortstack{42.1\\{\tiny{[40.6,43.6]}}}
        & \shortstack{44.9\\{\tiny{[43.3,46.4]}}}
        & \shortstack{48.6\\{\tiny{[46.9,50.2]}}}
        & \shortstack{$\mathbf{96.0}$\\{\tiny{[95.1,96.6]}}} \\[4pt]
        
        & $P_{\mathrm{punct}}$
        & \shortstack{31.5\\{\tiny{[30.4,32.5]}}}
        & \shortstack{28.8\\{\tiny{[27.8,29.9]}}}
        & \shortstack{33.6\\{\tiny{[33.3,35.8]}}}
        & \shortstack{27.2\\{\tiny{[26.3,28.1]}}}
        & \shortstack{$\mathbf{91.5}$\\{\tiny{[90.7,92.2]}}} \\[4pt]
        
        \bottomrule
    \end{tabular}
\end{table}

In the English real-time setting (Table \ref{tab:med_term_en_realtime}), Symphony reduces WER to $1.4\%$, compared with $17.7\%$ for OpenAI, $18.1\%$ for ElevenLabs, $17.4\%$ for Whisper, and $18.9\%$ for Parakeet. Symphony also obtains the highest medical term recall, reaching $84.1\%$, compared with $75.5\%$ for the strongest non-Corti system. These results indicate that Symphony substantially improves recognition of specialized medical terminology while also reducing overall transcription errors.

The largest differences appear on formatting and spoken-punctuation metrics. On English MedTerm, Symphony reaches $98.3\%$ recall on formatted entities, while the strongest baseline reaches only $44.3\%$ in the real-time comparison and $51.2\%$ in the offline comparison. Similarly, Symphony obtains $92.4\%$ spoken-punctuation recall and $70.3\%$ spoken-punctuation precision, substantially outperforming all baselines. This gap is important because formatted entities and spoken punctuation are central to clinical dictation: dosages, measurements, dates, and punctuation commands must be rendered correctly for transcripts to be useful in documentation workflows.

\begin{table}[t]
    \centering
    \caption{\textbf{Offline on MedTerm against real-time Symphony [English].} Comparison across WER and keyterm precision/recall ($P_{\mathcal{V}}, R_{\mathcal{V}}$) for medical entities ($\mathcal{V}=\mathrm{med}$), spoken-punctuation symbols ($\mathcal{V}=\mathrm{punct}$), and formatted entities ($\mathcal{V}=\mathrm{fmt}$). Higher is better for $P/R$ metrics; lower is better for WER. All values reported in percent $[\%]$. 95\% confidence intervals shown below each value. \textbf{Bold} indicates best in each row.}
    \vspace{0.5cm}
    \label{tab:med_term_en}
    \small
    \begin{tabular}{llcccccc}
        \toprule
        & & \multicolumn{4}{c}{\textbf{Closed-source}} & \textbf{Corti} \\
        \cmidrule(lr){3-6} \cmidrule(lr){7-7}
        \textbf{Lang} & \textbf{Metric}
        & \rotatebox{90}{\textbf{Google}}
        & \rotatebox{90}{\textbf{AWS}}
        & \rotatebox{90}{\textbf{OpenAI}}
        & \rotatebox{90}{\textbf{ElevenLabs}}
        & \rotatebox{90}{\textbf{Symphony}} \\
        \midrule
        & WER
        & \shortstack{18.6\\{\tiny{[17.9,19.2]}}}
        & \shortstack{16.4\\{\tiny{[15.9,16.8]}}}
        & \shortstack{12.8\\{\tiny{[12.2,13.4]}}}
        & \shortstack{17.4\\{\tiny{[16.8,18.0]}}}
        & \shortstack{$\mathbf{2.1}$\\{\tiny{[2.0,2.3]}}} \\[4pt]
        
        & $R_{\mathrm{med}}$
        & \shortstack{50.3\\{\tiny{[47.8,52.9]}}}
        & \shortstack{64.3\\{\tiny{[61.8,66.9]}}}
        & \shortstack{81.0\\{\tiny{[78.9,83.2]}}}
        & \shortstack{80.5\\{\tiny{[78.3,82.6]}}}
        & \shortstack{$\mathbf{81.8}$\\{\tiny{[80.0,83.7]}}} \\[4pt]
        
     EN & $R_{\mathrm{fmt}}$
        & \shortstack{45.8\\{\tiny{[42.5,49.3]}}}
        & \shortstack{44.2\\{\tiny{[41.4,47.2]}}}
        & \shortstack{51.2\\{\tiny{[47.6,54.4]}}}
        & \shortstack{33.7\\{\tiny{[29.9,37.4]}}}
        & \shortstack{$\mathbf{93.0}$\\{\tiny{[91.5,94.3]}}} \\[4pt]
        
        & $R_{\mathrm{punct}}$
        & \shortstack{33.7\\{\tiny{[32.3,35.2]}}}
        & \shortstack{43.9\\{\tiny{[42.5,45.4]}}}
        & \shortstack{48.7\\{\tiny{[46.8,50.6]}}}
        & \shortstack{45.7\\{\tiny{[44.1,47.2]}}}
        & \shortstack{$\mathbf{96.0}$\\{\tiny{[95.1,96.6]}}} \\[4pt]
        
        & $P_{\mathrm{punct}}$
        & \shortstack{46.9\\{\tiny{[44.8,48.8]}}}
        & \shortstack{31.6\\{\tiny{[30.6,32.6]}}}
        & \shortstack{39.4\\{\tiny{[37.7,41.0]}}}
        & \shortstack{32.0\\{\tiny{[30.8,33.0]}}}
        & \shortstack{$\mathbf{91.5}$\\{\tiny{[90.7,92.2]}}} \\[4pt]
        
        \bottomrule
    \end{tabular}
\end{table}

The offline comparison (Table \ref{tab:med_term_en}) shows that these gains are not only due to comparing against weaker real-time systems. Even when evaluating against stronger offline APIs, the real-time Symphony remains best on every metric. OpenAI is the strongest non-Corti baseline on English MedTerm, with $12.8\%$ WER, $81.0\%$ medical term recall, and $51.2\%$ formatted entity recall. Symphony improves over this with $1.4\%$ WER, $84.1\%$ medical term recall, and $98.3\%$ formatted entity recall. The difference is especially pronounced for formatting and spoken punctuation, suggesting that Symphony is not merely a stronger recognizer, but better aligned with the output requirements of medical dictation.

\begin{table}[t]
    \centering
    \caption{\textbf{Real-time on MedTerm [German, French].} Comparison across WER and keyterm precision/recall ($P_{\mathcal{V}}, R_{\mathcal{V}}$) for medical entities ($\mathcal{V}=\mathrm{med}$), spoken-punctuation symbols ($\mathcal{V}=\mathrm{punct}$), and formatted entities ($\mathcal{V}=\mathrm{fmt}$). Higher is better for $P/R$ metrics; lower is better for WER. All values reported in percent $[\%]$. 95\% confidence intervals shown below each value. \textbf{Bold} indicates best in each row.}
    \vspace{0.5cm}
    \label{tab:med_term_fr_de_realtime}
    \small
    \begin{tabular}{llccccc}
        \toprule
        & & \multicolumn{2}{c}{\textbf{Open-source}} & \multicolumn{2}{c}{\textbf{Closed-source}} & \textbf{Corti} \\
        \cmidrule(lr){3-4} \cmidrule(lr){5-6} \cmidrule(lr){7-7}
        \textbf{Lang} & \textbf{Metric}
        & \rotatebox{90}{\textbf{Parakeet}}
        & \rotatebox{90}{\textbf{Whisper}}
        & \rotatebox{90}{\textbf{OpenAI}}
        & \rotatebox{90}{\textbf{ElevenLabs}}
        & \rotatebox{90}{\textbf{Symphony}} \\
        \midrule
        & WER
        & \shortstack{15.9\\{\tiny{[15.5,16.4]}}}
        & \shortstack{15.6\\{\tiny{[15.1,16.0]}}}
        & \shortstack{13.0\\{\tiny{[12.6,13.4]}}}
        & \shortstack{16.4\\{\tiny{[15.9,17.0]}}}
        & \shortstack{$\mathbf{2.5}$\\{\tiny{[2.3,2.7]}}} \\[4pt]
        
        & $R_{\mathrm{med}}$
        & \shortstack{51.1\\{\tiny{[48.6,53.7]}}}
        & \shortstack{57.1\\{\tiny{[54.8,59.7]}}}
        & \shortstack{70.6\\{\tiny{[68.0,73.0]}}}
        & \shortstack{62.9\\{\tiny{[60.3,65.5]}}}
        & \shortstack{$\mathbf{75.6}$\\{\tiny{[73.4,77.9]}}} \\[4pt]
        
     DE & $R_{\mathrm{fmt}}$
        & \shortstack{81.3\\{\tiny{[79.5,83.0]}}}
        & \shortstack{81.9\\{\tiny{[80.2,83.7]}}}
        & \shortstack{84.8\\{\tiny{[83.2,86.3]}}}
        & \shortstack{84.5\\{\tiny{[82.9,85.9]}}}
        & \shortstack{$\mathbf{99.5}$\\{\tiny{[99.3,99.8]}}} \\[4pt]
        
        & $R_{\mathrm{punct}}$
        & \shortstack{32.4\\{\tiny{[30.0,34.9]}}}
        & \shortstack{33.9\\{\tiny{[31.4,36.5]}}}
        & \shortstack{40.3\\{\tiny{[37.6,43.2]}}}
        & \shortstack{38.8\\{\tiny{[36.3,41.6]}}}
        & \shortstack{$\mathbf{94.1}$\\{\tiny{[93.2,95.2]}}} \\[4pt]
        
        & $P_{\mathrm{punct}}$
        & \shortstack{20.0\\{\tiny{[18.3,21.9]}}}
        & \shortstack{20.1\\{\tiny{[18.6,21.9]}}}
        & \shortstack{25.6\\{\tiny{[23.6,27.8]}}}
        & \shortstack{19.2\\{\tiny{[17.8,20.8]}}}
        & \shortstack{$\mathbf{90.8}$\\{\tiny{[89.3,92.1]}}} \\[4pt]
        
        \cmidrule(l){2-7}
        
        & WER
        & \shortstack{14.7\\{\tiny{[14.2,15.2]}}}
        & \shortstack{13.9\\{\tiny{[13.5,14.4]}}}
        & \shortstack{10.6\\{\tiny{[10.1,11.0]}}}
        & \shortstack{16.0\\{\tiny{[15.5,16.6]}}}
        & \shortstack{$\mathbf{2.9}$\\{\tiny{[2.7,3.1]}}} \\[4pt]
        
        & $R_{\mathrm{med}}$
        & \shortstack{60.5\\{\tiny{[57.9,63.1]}}}
        & \shortstack{64.3\\{\tiny{[61.6,67.0]}}}
        & \shortstack{73.9\\{\tiny{[71.5,76.2]}}}
        & \shortstack{69.9\\{\tiny{[67.3,72.6]}}}
        & \shortstack{$\mathbf{80.2}$\\{\tiny{[78.1,82.3]}}} \\[4pt]
        
     FR & $R_{\mathrm{fmt}}$
        & \shortstack{62.7\\{\tiny{[60.2,65.2]}}}
        & \shortstack{64.0\\{\tiny{[61.4,66.8]}}}
        & \shortstack{69.6\\{\tiny{[67.1,72.1]}}}
        & \shortstack{58.0\\{\tiny{[55.0,61.0]}}}
        & \shortstack{$\mathbf{93.5}$\\{\tiny{[92.2,94.8]}}} \\[4pt]
        
        & $R_{\mathrm{punct}}$
        & \shortstack{28.7\\{\tiny{[27.7,29.7]}}}
        & \shortstack{27.0\\{\tiny{[25.9,28.0]}}}
        & \shortstack{33.2\\{\tiny{[31.5,34.8]}}}
        & \shortstack{33.4\\{\tiny{[32.2,34.7]}}}
        & \shortstack{$\mathbf{91.3}$\\{\tiny{[90.3,92.4]}}} \\[4pt]
        
        & $P_{\mathrm{punct}}$
        & \shortstack{23.6\\{\tiny{[22.6,24.6]}}}
        & \shortstack{21.5\\{\tiny{[20.4,22.3]}}}
        & \shortstack{34.9\\{\tiny{[32.9,36.8]}}}
        & \shortstack{24.0\\{\tiny{[23.0,25.1]}}}
        & \shortstack{$\mathbf{92.3}$\\{\tiny{[91.5,93.0]}}} \\[4pt]
        
        \bottomrule
    \end{tabular}
\end{table}

The German and French results (Table \ref{tab:med_term_fr_de_realtime}) show that the same pattern holds beyond English. For German, Symphony achieves $2.4\%$ WER, compared with $13.0\%$ for OpenAI, $16.4\%$ for ElevenLabs, $15.6\%$ for Whisper, and $15.9\%$ for Parakeet. Symphony also obtains the best medical term recall at $76.4\%$, the best formatted entity recall at $99.4\%$, and the best spoken-punctuation recall at $96.1\%$. For French, Symphony achieves $3.9\%$ WER, compared with $10.6\%$ for OpenAI, $16.0\%$ for ElevenLabs, $13.9\%$ for Whisper, and $14.7\%$ for Parakeet. It also reaches the highest medical term recall, formatted entity recall, spoken-punctuation recall, and spoken-punctuation precision.

Overall, the MedTerm results demonstrate that Symphony provides large improvements over both open-source systems and leading closed-source speech-to-text APIs. The improvements are consistent across English, German, and French, and are strongest on the clinically important categories that general-purpose systems handle poorly: medical terms, formatted entities, and spoken punctuation. This supports the central design goal of Symphony: producing transcripts that are not only acoustically accurate, but also formatted and structured for real-world medical dictation workflows.

\subsection{Medical dictation}

\begin{table}[t]
    \centering
    \caption{\textbf{Real-time on MedDictate [English].} Comparison across WER and keyterm precision/recall ($P_{\mathcal{V}}, R_{\mathcal{V}}$) for medical entities ($\mathcal{V}=\mathrm{med}$), spoken-punctuation symbols ($\mathcal{V}=\mathrm{punct}$), and formatted entities ($\mathcal{V}=\mathrm{fmt}$). Higher is better for $P/R$ metrics; lower is better for WER. All values reported in percent $[\%]$. 95\% confidence intervals shown below each value. \textbf{Bold} indicates best in each row.}
    \vspace{0.5cm}
    \label{tab:med_dictate_en}
    \small
    \begin{tabular}{llcccc}
        \toprule
        & & \multicolumn{2}{c}{\textbf{Open-source}} & \textbf{Closed-source} & \textbf{Corti} \\
        \cmidrule(lr){3-4} \cmidrule(lr){5-5} \cmidrule(lr){6-6}
        \textbf{Lang} & \textbf{Metric}
        & \rotatebox{90}{\textbf{Parakeet}}
        & \rotatebox{90}{\textbf{Whisper}}
        & \rotatebox{90}{\textbf{DMO}}
        & \rotatebox{90}{\textbf{Symphony}} \\
        \midrule
        & WER
        & \shortstack{48.3\\{\tiny{[39.9,56.9]}}}
        & \shortstack{40.3\\{\tiny{[34.2,47.0]}}}
        & \shortstack{5.7\\{\tiny{[4.1,7.3]}}}
        & \shortstack{$\mathbf{4.1}$\\{\tiny{[3.1,5.3]}}} \\[4pt]
        
        & $R_{\mathrm{med}}$
        & \shortstack{81.4\\{\tiny{[77.2,85.6]}}}
        & \shortstack{90.0\\{\tiny{[87.4,92.5]}}}
        & \shortstack{92.9\\{\tiny{[90.1,95.5]}}}
        & \shortstack{$\mathbf{93.9}$\\{\tiny{[91.9,95.5]}}} \\[4pt]
        
     EN & $R_{\mathrm{fmt}}$
        & \shortstack{14.9\\{\tiny{[10.8,20.8]}}}
        & \shortstack{37.8\\{\tiny{[32.4,47.8]}}}
        & \shortstack{$\mathbf{78.3}$\\{\tiny{[68.8,92.2]}}}
        & \shortstack{72.9\\{\tiny{[65.0,81.7]}}} \\[4pt]
        
        & $R_{\mathrm{punct}}$
        & \shortstack{41.3\\{\tiny{[36.4,46.4]}}}
        & \shortstack{38.9\\{\tiny{[34.4,43.2]}}}
        & \shortstack{94.9\\{\tiny{[93.1,96.6]}}}
        & \shortstack{$\mathbf{95.2}$\\{\tiny{[93.4,97.0]}}} \\[4pt]
        
        & $P_{\mathrm{punct}}$
        & \shortstack{37.5\\{\tiny{[32.1,43.0]}}}
        & \shortstack{34.2\\{\tiny{[29.0,39.6]}}}
        & \shortstack{97.4\\{\tiny{[96.1,98.6]}}}
        & \shortstack{$\mathbf{98.1}$\\{\tiny{[97.2,99.0]}}} \\[4pt]
        
        \bottomrule
    \end{tabular}
\end{table}

Table~\ref{tab:med_dictate_en} evaluates Symphony on MedDictate, a realistic English medical dictation benchmark, against both general open-source ASR models and DMO, a front-end speech recognition system specialized for clinical dictation. This comparison is important because DMO represents a strong and widely adopted medical dictation baseline, whereas Parakeet and Whisper provide a sanity check for how far general-purpose open-source models are from meeting the requirements of this setting.

The open-source systems perform poorly on MedDictate. Parakeet and Whisper obtain WERs of $48.3\%$ and $40.3\%$, respectively, indicating that general-purpose ASR models struggle with realistic medical dictation despite reasonable medical term recall. Their performance is especially weak on formatted entities and spoken punctuation: Parakeet reaches only $14.9\%$ formatted entity recall and Whisper reaches $37.8\%$, while both systems remain below $42\%$ spoken-punctuation recall and below $38\%$ spoken-punctuation precision. These results show that medical dictation is not only a recognition problem, but also requires domain-specific handling of formatting, punctuation commands, and clinically meaningful surface forms.

Compared with DMO, Symphony achieves a lower WER ($4.6\%$ vs.\ $5.7\%$), slightly higher medical term recall ($93.5\%$ vs.\ $92.9\%$), and comparable spoken-punctuation precision — reaching the performance regime of a dedicated medical dictation system while being exposed through an API designed for broader clinical workflows. DMO remains strongest on formatted entity recall and spoken-punctuation recall, reaching $78.3\%$ and $94.9\%$ compared with Symphony's $74.4\%$ and $89.0\%$. However, DMO's medical-term false discovery rate is almost twice that of Symphony ($1.33\%$ vs.\ $0.79\%$), evidence that its strong recall partly reflects over-aggressive vocabulary priors that insert medical terms where none were spoken. This over-specialization tendency is also visible in the qualitative comparison on conversational speech in Figure~\ref{fig:dmo_generalization_example}.

\subsection{General-domain speech recognition}

\begin{table}[t]
    \centering
    \caption{\textbf{Real-time on CommonVoice.} General-domain speech recognition comparison across English (EN), German (DE), and French (FR). We report normalized word error rate (WER). Lower is better. All values reported in percent $[\%]$. 95\% confidence intervals shown below each value. \textbf{Bold} indicates best in each row.}
    \vspace{0.5cm}
    \label{tab:commonvoice_realtime}
    \small
    \begin{tabular}{llccccc}
        \toprule
        & & \multicolumn{2}{c}{\textbf{Open-source}} & \multicolumn{2}{c}{\textbf{Closed-source}} & \textbf{Corti} \\
        \cmidrule(lr){3-4} \cmidrule(lr){5-6} \cmidrule(lr){7-7}
        \textbf{Lang} & \textbf{Metric}
        & \rotatebox{90}{\textbf{Parakeet}}
        & \rotatebox{90}{\textbf{Whisper}}
        & \rotatebox{90}{\textbf{OpenAI}}
        & \rotatebox{90}{\textbf{ElevenLabs}}
        & \rotatebox{90}{\textbf{Symphony}} \\
        \midrule
     EN & WER
        & \shortstack{11.7\\{\tiny{[10.8,12.6]}}}
        & \shortstack{13.1\\{\tiny{[12.1,14.1]}}}
        & \shortstack{13.3\\{\tiny{[12.4,14.3]}}}
        & \shortstack{11.3\\{\tiny{[10.5,12.2]}}}
        & \shortstack{$\mathbf{11.2}$\\{\tiny{[10.3,12.1]}}} \\[4pt]
        
        \cmidrule(l){2-7}
        
     DE & WER
        & \shortstack{7.2\\{\tiny{[6.5,7.9]}}}
        & \shortstack{7.9\\{\tiny{[7.2,8.6]}}}
        & \shortstack{6.7\\{\tiny{[6.1,7.3]}}}
        & \shortstack{6.9\\{\tiny{[6.3,7.6]}}}
        & \shortstack{$\mathbf{6.6}$\\{\tiny{[5.9,7.3]}}} \\[4pt]
        
        \cmidrule(l){2-7}
        
     FR & WER
        & \shortstack{8.2\\{\tiny{[7.5,8.9]}}}
        & \shortstack{11.4\\{\tiny{[10.5,12.3]}}}
        & \shortstack{10.5\\{\tiny{[9.7,11.2]}}}
        & \shortstack{10.5\\{\tiny{[9.7,11.3]}}}
        & \shortstack{$\mathbf{8.0}$\\{\tiny{[7.3,8.7]}}} \\[4pt]
        
        \bottomrule
    \end{tabular}
\end{table}

To verify that the medical-domain gains do not come at the expense of general speech recognition quality, we evaluate all real-time systems on CommonVoice in English, German, and French. This benchmark serves as a general-domain control setting: unlike MedTerm and MedDictate, it does not specifically target clinical terminology, formatting, or dictation commands.

Table~\ref{tab:commonvoice_realtime} shows that Symphony remains competitive with strong general-purpose systems across all three languages. In English, Symphony obtains the lowest WER at $11.2\%$, closely followed by ElevenLabs at $11.3\%$, also outperforming Whisper and OpenAI real-time. In German, Symphony also achieves the best WER at $6.6\%$, slightly ahead of OpenAI real-time at $6.7\%$ and ElevenLabs real-time at $6.9\%$. In French, Symphony again obtains the lowest WER at $8.0\%$, outperforming Parakeet, OpenAI realtime, ElevenLabs realtime, and Whisper.

These results indicate that Symphony's specialization for medical speech does not lead to a loss of general-domain recognition performance. Instead, Symphony remains on par with or better than leading real-time baselines on CommonVoice, while substantially outperforming the same systems on medical terminology, formatting, and dictation-oriented metrics. This supports the conclusion that Symphony's improvements in clinical settings reflect domain-specific capability rather than overfitting to narrow medical evaluation data.

\begin{figure}[p]
    \centering
    \captionsetup[subfigure]{skip=-1ex}
    \begin{subfigure}[t]{\linewidth}
        \centering
        \includegraphics[width=\linewidth]{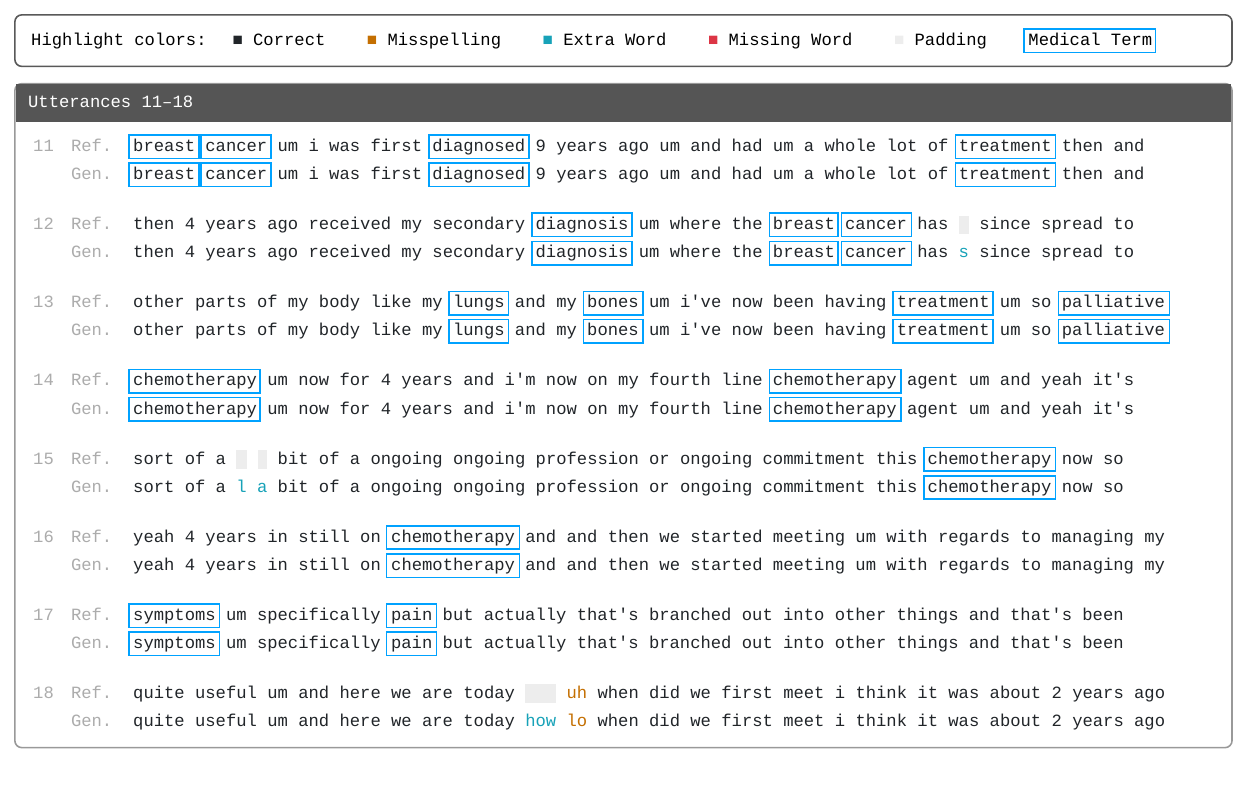}
        \caption{Symphony}
        \label{fig:generalization_corti}
    \end{subfigure}
    
    \vspace{1em}
    
    \begin{subfigure}[t]{\linewidth}
        \centering
        \includegraphics[width=\linewidth]{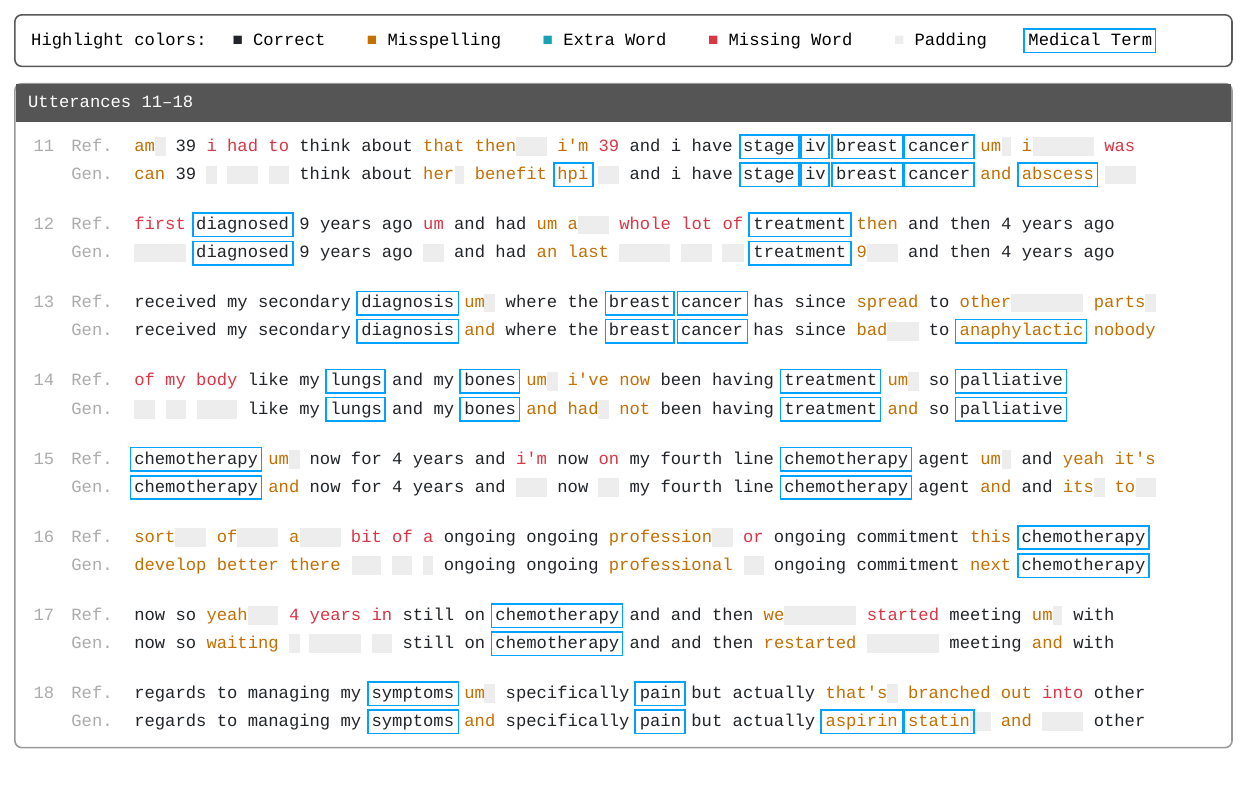}
        \caption{Dragon Medical One}
        \label{fig:generalization_dmo}
    \end{subfigure}
    \caption{Qualitative comparison of Symphony and Dragon Medical One (DMO) on non-dictation conversational speech, visualized with \href{https://pypi.org/project/corti-canal/}{Corti-Canal}. Both systems transcribe the same audio. DMO, designed for single-speaker front-end dictation, occasionally hallucinates medical terms for non-medical ones and generalizes less reliably to conversational settings. Reference alignments differ slightly between (\subref{fig:generalization_corti}) and (\subref{fig:generalization_dmo}) due to preceding context.}
    \label{fig:dmo_generalization_example}
\end{figure}

On the other hand, Google Medical ASR and DMO illustrate the risks of narrow domain specialization. In our evaluations, Google Medical reached 40.0\% WER on English CommonVoice, far worse than any real-time system in Table~\ref{tab:commonvoice_realtime} despite being offline itself, indicating poor generalization to non-clinical speech. DMO could not be systematically evaluated on CommonVoice, as it is a front-end dictation application, but as noted above, its elevated false discovery rate for medical terms on MedDictate and qualitative examples of conversational speech (Figure~\ref{fig:dmo_generalization_example}) show the same signs of over-specialization; the system inserts medical terms where none were spoken, even in medical audio.

\subsection{Keyterm biasing}
We evaluate keyterm biasing by providing Symphony with the set of medical terms present in each evaluation dataset and measuring its effect on medical term false negatives. As shown in Table~\ref{tab:keyterm_biasing}, biasing substantially improves medical term coverage. On MedDictate, the medical term false-negative rate is reduced by $47.9\%$ relative to the unbiased baseline, while on MedRad the reduction is $50.9\%$. This indicates that keyterm biasing is effective for improving recognition of clinically important terminology that may otherwise be missed.

Importantly, the improvement in recall does not come at the cost of precision. Medical term precision remains essentially unchanged under biasing, rising slightly from $98.5\%$ to $98.7\%$ on MedDictate and decreasing only marginally from $97.0\%$ to $96.7\%$ on MedRad. This suggests that Symphony can leverage user- or workflow-specific terminology to recover more relevant medical terms without introducing spurious medical entities. In clinical applications, this is critical: keyterm biasing should improve recognition of expected terminology—medications, procedures, or site-specific vocabulary—without hallucinating terms that could alter the clinical meaning of the transcript.

\begin{table}[t]
    \centering
    \caption{\textbf{Keyterm biasing evaluation.} Effect of biasing Symphony
    toward the full set of medical terms present in each dataset.
    $\mathrm{rFNR}_{\mathrm{med}}$ is the relative reduction in
    false-negative rate compared with the unbiased baseline (higher is
    better) for medical terms ($\mathcal{V} = \mathrm{med})$; $P_{\mathrm{med}}$ is reported under both conditions to
    verify that biasing does not degrade precision.}\vspace{0.25cm}
    \label{tab:keyterm_biasing}
    \begin{tabular}{lccc}
        \toprule
        Dataset
        & $\mathrm{rFNR}_{\mathrm{med}}$
        & $P_{\mathrm{med}}^{\mathrm{unbiased}}$
        & $P_{\mathrm{med}}^{\mathrm{biased}}$ \\
        \midrule
        MedDictate (EN)  & 47.9\% & 98.5\% & 98.7\% \\
        MedRad (EN)     & 50.9\% & 97.0\% & 96.7\% \\
        \bottomrule
    \end{tabular}
\end{table}

\begin{figure}[!b]
    \centering
    \includegraphics[width=0.68\linewidth]{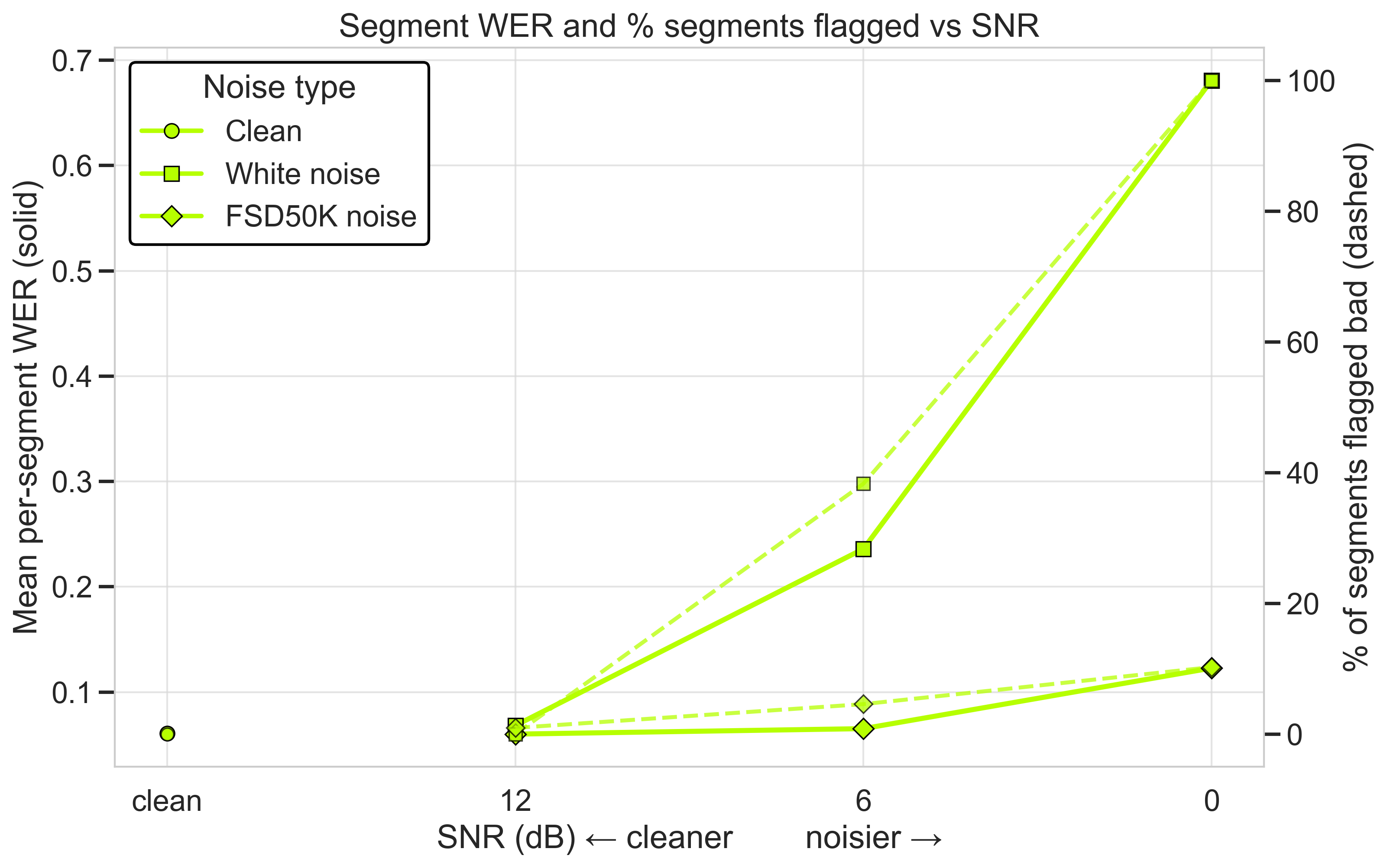}
    \caption{Audio-quality event detection compared with transcription degradation under increasing noise. Solid lines show mean per-segment WER, while dashed lines show the percentage of segments flagged as bad audio. As signal-to-noise ratio decreases, both WER and the rate of audio-quality events increase, indicating that Symphony surfaces degraded input conditions when associated with worse transcription quality.}
    \label{fig:audio_quality_events}
\end{figure}

\subsection{Audio quality events}

In addition to transcription accuracy, Symphony exposes real-time audio-quality events that can alert applications when input conditions are likely to degrade recognition performance. We evaluate these events by adding different types of noise to speech segments and comparing the percentage of segments flagged as low quality against the resulting per-segment WER.

Figure~\ref{fig:audio_quality_events} shows that the audio-quality events track the degradation in transcription quality as the signal-to-noise ratio decreases. For clean audio, the system flags few or no segments, and WER remains low. As the audio becomes noisier, especially under white-noise corruption, both the mean per-segment WER and the percentage of flagged segments increase sharply. This indicates that the health events are not arbitrary, but correlate with the conditions where recognition quality deteriorates.

These results support the use of audio-quality events as an operational signal in real-time clinical voice interfaces. Rather than discovering input-quality problems only after a transcript has been produced, applications can surface warnings during the interaction itself, prompting users to adjust microphone placement, reduce background noise, or repeat an utterance. This makes audio-quality monitoring an important complement to model accuracy in safety-critical speech workflows.

\section{Conclusion}

We presented Symphony for Speech-to-Text, a medical-grade speech recognition system designed for real-time streaming and batch file-based clinical workflows. Symphony combines medical speech recognition, structured formatting, contextual correction, command support, audio-quality signals, and keyterm biasing within a unified API for dictation, conversational transcription, and batch processing.

Across medical terminology and medical dictation evaluations, Symphony substantially improves over strong open-source and closed-source baselines on the metrics most important for clinical use: WER, medical term recall, formatted entity recall, and spoken-punctuation handling. At the same time, results on CommonVoice show that this specialization does not come at the cost of general-domain robustness. Compared with dedicated front-end dictation systems, Symphony reaches the same performance regime while exposing a more flexible infrastructure for speech-enabled clinical applications.

These results support the view that medical speech recognition should not be treated as a single transcription task. Reliable clinical voice interfaces require recognition, formatting, contextual correction, and workflow adaptation to work together. Symphony demonstrates that such an integrated design can improve both transcription accuracy and clinical usability, providing a foundation for future speech-enabled healthcare systems.

\section*{Acknowledgments}

We thank Michael Thorsager, Simas Paulikas, Jerome Amon, Gediminas Kikilas, Chituru Chinwah, Kevin Pelgrims, Nicklas Frahm and the Corti AI Platform team for their invaluable support in the build, deploy, and maintenance of the microservice architecture that made this work possible. Thank you to Dr. Lasse Krogsbøll, Dr. Vanessa Klungtvedt, Dr. Marvin Dumke, James Kane, Majed Sharif, and Henrik Cullen for their contributions, insightful discussions, and support throughout the completion of this work.

\newpage
\bibliography{references}

\newpage
\appendix

\section{Configuration of benchmarked systems}
\label{app:asr-configs}

This section documents the inference-time configuration for the four
external ASR systems evaluated in this work.
AWS Transcribe Medical and Google Cloud Speech-to-Text are batch REST APIs
with dedicated medical model tiers; audio is routed via cloud storage for
longer recordings.
ElevenLabs and OpenAI Realtime are WebSocket streaming APIs that simulate
live dictation: audio is pushed in 100~ms frames at real-time pace and
transcripts are emitted incrementally within a single long-lived session.
No vocabulary biasing, prompt injection, or output post-processing was
applied to any system.
Table~\ref{tab:asr-configs} summarises the key parameters.

\begin{table}[!h]
    \centering
    \small
    \caption{Inference-time configuration for the four closed-source ASR systems. Entries marked ``---'' indicate the parameter does not apply.}\vspace{0.25cm}
    \label{tab:asr-configs}
    \begin{threeparttable}
        \resizebox{\textwidth}{!}{%
        \begin{tabular}{lllll}
            \toprule
            \textbf{Parameter}
            & \textbf{AWS Medical}
            & \textbf{Google Medical}
            & \textbf{ElevenLabs}
            & \textbf{OpenAI Realtime} \\
            \midrule
            Model / tier
            & Medical DICTATION
            & \texttt{medical\_dictation}
            & \texttt{scribe\_v2\_realtime}
            & \texttt{gpt-realtime-whisper} \\
            Interface
            & REST (async)
            & REST (sync / async)
            & WebSocket
            & WebSocket \\
            Language
            & en-US
            & en-US
            & configurable
            & configurable \\
            Sample rate
            & 16{,}000~Hz
            & 16{,}000~Hz
            & 16{,}000~Hz
            & 24{,}000~Hz\textsuperscript{a} \\
            Audio routing
            & S3 upload
            & inline / GCS\textsuperscript{b}
            & streamed frames
            & streamed frames \\
            Automatic punctuation
            & native
            & enabled
            & enabled
            & native \\
            \midrule
            \multicolumn{5}{l}{\textit{Streaming parameters}} \\
            Frame size        & ---   & ---   & 100~ms        & 100~ms \\
            Real-time factor  & ---   & ---   & $1\times$     & $1\times$ \\
            Commit strategy   & ---   & ---   & server VAD\textsuperscript{c}    & manual \\
            VAD silence threshold & --- & ---  & 2.5~s         & --- \\
            VAD threshold     & ---   & ---   & 0.5           & --- \\
            Post-send grace   & ---   & ---   & 5.0~s         & 5.0~s \\
            \midrule
            \multicolumn{5}{l}{\textit{Request settings}} \\
            Parallel requests & 4     & 4     & 6             & 6 \\
            Max retries       & ---   & 3     & 3             & 3 \\
            Initial backoff   & ---   & 1~s   & 1~s           & 1~s \\
            Max backoff       & ---   & 10~s  & 10~s          & 10~s \\
            Jitter            & ---   & $\pm$0.5~s & $\pm$0.5~s & $\pm$0.5~s \\
            \bottomrule
        \end{tabular}%
        }
        \begin{tablenotes}
        \tiny
        \item[a] Input audio at lower sample rates (e.g.\ 16{,}000~Hz) is automatically resampled to 24{,}000~Hz before transmission, as required by the API.
        \item[b] Audio shorter than 60~s is sent inline via \texttt{recognize}; longer audio is uploaded to GCS and processed via \texttt{longRunningRecognize}.
        \item[c] For the Common Voice dataset, the \texttt{manual} commit strategy was used instead.
        \end{tablenotes}
    \end{threeparttable}
\end{table}

\begin{table}[!h]
    \centering
    \small
    \caption{Inference-time configuration for the two open-source ASR systems. To effectively perform streamed speech recognition, both systems receive audio segments from an external voice activity detector. Entries marked ``---'' indicate the parameter does not apply.}\vspace{0.25cm}
    \label{tab:asr-configs-opensource}
    \begin{threeparttable}
        \begin{tabular}{lll}
            \toprule
            \textbf{Parameter}
            & \textbf{Whisper}
            & \textbf{Parakeet} \\
            \midrule
            Model
            & \texttt{whisper-large-v3}
            & \texttt{parakeet-tdt-0.6b}\textsuperscript{a} \\
            Framework
            & \texttt{faster-whisper}
            & NVIDIA NeMo \\
            Language
            & set per dataset
            & auto-detect (v3) / EN-only (v2) \\
            Sample rate
            & 16{,}000~Hz
            & 16{,}000~Hz \\
            Decoding
            & beam search (width 5)
            & greedy \\
            Temperature fallback
            & disabled
            & --- \\
            \midrule
            \multicolumn{3}{l}{\textit{Pseudo-streaming parameters}} \\
            Segmentation
            & external VAD
            & external VAD \\
            Segment duration
            & 2--12~s (median 8~s)
            & 2--12~s (median 8~s) \\
            \bottomrule
        \end{tabular}
        \begin{tablenotes}
        \tiny
        \item[a] English evaluations use Parakeet TDT v2 as this outperformed the multilingual v3; German and French evaluations used Parakeet TDT v3, which supports automatic language detection.
        \end{tablenotes}
    \end{threeparttable}
\end{table}

\end{document}